\documentclass[english,conference]{IEEEtran}
\usepackage[T1]{fontenc}
\usepackage[latin9]{inputenc}
\usepackage{color}
\usepackage{babel}
\usepackage{array}
\usepackage{float}
\usepackage{multirow}
\usepackage{algorithm2e}
\usepackage{amstext}
\usepackage{amssymb}
\usepackage{graphicx}
\PassOptionsToPackage{normalem}{ulem}
\usepackage{ulem}
\usepackage[pdfusetitle,
 bookmarks=false,
 breaklinks=false,pdfborder={0 0 1},backref=false,colorlinks=true]
 {hyperref}

\makeatletter

\providecommand{\tabularnewline}{\\}

\usepackage{cite}
\usepackage{algorithmic}
\usepackage{graphicx}
\usepackage{textcomp}
\usepackage{xcolor}
\usepackage{balance}

\usepackage{colortbl} 
\definecolor{header_color}{rgb}{0.74,0.88,0.91}
\definecolor{even_color}{rgb}{0.9,0.9,0.9}
\definecolor{subheader_color}{rgb}{0.85,0.93,0.95}
\definecolor{childheader_color}{rgb}{1.0,0.93,0.87}

\ifdefined\showcaptionsetup
 \PassOptionsToPackage{caption=false}{subfig}
\fi
\usepackage{subfig}
\makeatother

\begin{document}
\IEEEoverridecommandlockouts 
\def\BibTeX{{\rm B\kern-.05em{\sc i\kern-.025em b}\kern-.08em     
T\kern-.1667em\lower.7ex\hbox{E}\kern-.125emX}} 
\title{Large Language Models for Imbalanced Classification: Diversity makes
the difference}
\author{Dang Nguyen\textsuperscript{1}, Sunil Gupta\textsuperscript{1},
Kien Do\textsuperscript{1}, Thin Nguyen\textsuperscript{1}, Taylor
Braund\textsuperscript{2}, Alexis Whitton\textsuperscript{2}, Svetha
Venkatesh\textsuperscript{1}\\
\textsuperscript{1}Applied Artificial Intelligence Initiative (A\textsuperscript{2}I\textsuperscript{2}),
Deakin University, Geelong, Australia\\
\textsuperscript{2}Black Dog Institute, University of New South Wales,
Sydney, Australia\\
\textit{\textsuperscript{\textit{1}}\{d.nguyen, sunil.gupta, k.do,
thin.nguyen, svetha.venkatesh\}@deakin.edu.au}\\
\textit{\textsuperscript{\textit{2}}\{t.braund, a.whitton\}@blackdog.org.au}}
\maketitle
\begin{abstract}
Oversampling is one of the most widely used approaches for addressing
imbalanced classification. The core idea is to generate additional
minority samples to rebalance the dataset. Most existing methods,
such as SMOTE, require converting categorical variables into numerical
vectors, which often leads to information loss. Recently, large language
model (LLM)--based methods have been introduced to overcome this
limitation. However, current LLM-based approaches typically generate
minority samples with \textit{limited diversity}, reducing robustness
and generalizability in downstream classification tasks.

To address this gap, we propose a novel LLM-based oversampling method
designed to enhance diversity. First, we introduce a sampling strategy
that conditions synthetic sample generation on both minority labels
and features. Second, we develop a new permutation strategy for fine-tuning
pre-trained LLMs. Third, we fine-tune the LLM not only on minority
samples but also on interpolated samples to further enrich variability.

Extensive experiments on 10 tabular datasets demonstrate that our
method significantly outperforms eight SOTA baselines. The generated
synthetic samples are both realistic and diverse. Moreover, we provide
theoretical analysis through an entropy-based perspective, proving
that our method encourages diversity in the generated samples.
\end{abstract}

\section{Introduction\label{sec:Introduction}}

Recently, applying modern machine learning (ML) models such as deep
learning and LLMs to tabular data has gained a significant attention
from both research communities and industrial companies \cite{Beltagy2019,Assefa2020,borisov2022deep,shwartz2022tabular}.
For example, \textit{Prior Labs} -- an AI start-up founded in late
2024 focuses on developing LLM-based solutions for tabular data and
has raised \$9.3 million from investors\footnote{https://fortune.com/2025/02/05/prior-labs-9-million-euro-preseed-funding-tabular-data-ai}.

However, there are many challenges when dealing with tabular data.
One of them is the \textit{imbalance} problem i.e. some classes have
much fewer samples than the others. For example, when dealing with
a medical dataset, a disease may be very rare and only affect a small
proportion of the population, leading to an imbalanced dataset. This
can lead to a biased classifier if trained on the imbalanced dataset,
where the classifier struggles to properly learn the characteristics
of the minority class and it favors the majority labels in its predictions
\cite{chawla2002smote,he2009learning,yang2024language}.

\begin{figure}[th]
\begin{centering}
\includegraphics[scale=0.48]{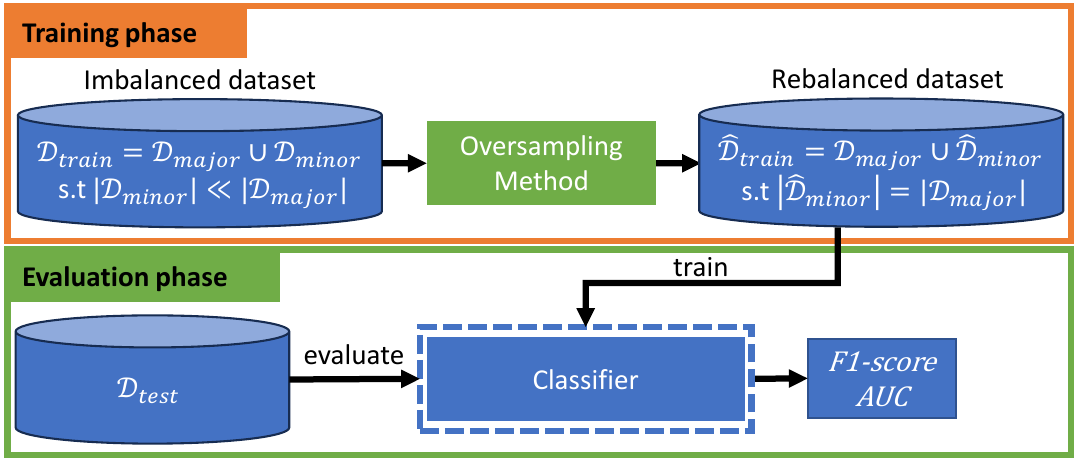}
\par\end{centering}
\caption{\label{fig:Training-and-evaluation}Training and evaluation of an
oversampling method. \textbf{Training:} the oversampling method learns
from the imbalanced dataset ${\cal D}_{train}$ to generate a synthetic
minority dataset $\hat{{\cal D}}_{minor}$ to create the rebalanced
dataset $\hat{{\cal D}}_{train}$. \textbf{Evaluation:} $\hat{{\cal D}}_{train}$
is used to train a ML classifier (e.g. XGBoost). The classifier is
evaluated on a held-out test set ${\cal D}_{test}$ to compute F1-score
and AUC. \textit{A better score implies a better oversampling method}.}
\end{figure}

One of the most popular solutions for the imbalance problem is \textit{oversampling},
which generates more minority samples to rebalance the dataset \cite{chawla2002smote,he2008adasyn,zhang2023generative,yang2024language}.
To evaluate an oversampling method, we often train a ML classifier
on the rebalanced dataset and compute performance scores on a test
set. A better score indicates a better oversampling method. Figure
\ref{fig:Training-and-evaluation} illustrates the training and evaluation
phases.

Most traditional oversampling methods based on SMOTE \cite{chawla2002smote}
require the conversion of categorical variables to numeric vectors
using ordinal or one-hot encoding. This pre-processing step may cause
information loss and artifact introduction \cite{Borisov2023}. LLM-based
methods for oversampling can overcome this problem as they represent
tabular data as text instead of numerical. Moreover, they can capture
the variable contexts e.g. the relationship between ``\textit{Gender}''
and ``\textit{Job}''. To generate a minority sample $\hat{x}$ with
$M$ features $\{X_{1},...,X_{M}\}$, existing LLM-based methods construct
a prompt conditioned on the minority label. Namely, they feed a prompt
``$Y$ is $y_{minor}$'' ($Y$ is the target variable and $y_{minor}$
is the minority label) as the initial tokens for LLMs to query the
next token. They then iteratively concatenate the predicted token
to the initial prompt and query the LLMs until all the features are
generated \cite{zhang2023generative,nguyen2024tabular}. This process
is straightforward as LLM-based methods support \textit{arbitrary
conditioning} i.e. the capacity to generate data conditioned on an
arbitrary set of variables. However, LLM-based methods often do not
generate diverse and generalizable minority samples, which is very
important for training classifiers in downstream tasks (see Figure
\ref{fig:Training-and-evaluation}).

To address this problem, we propose an LLM-based method with three
contributions. First, we observe that existing LLM-based methods construct
prompts based on the minority label solely \cite{Borisov2023,zhang2023generative,nguyen2024tabular}.
Although this is simple and intuitive, it generates less diverse synthetic
minority samples due to the softmax function used to predict the next
token. We propose a simple yet effective trick to fix this issue,
which \textit{forms the prompts based on both minority label and features}.
Second, the current LLM-based methods permute both the features and
the target variable in the fine-tuning data \cite{Borisov2023,zhang2023generative}.
As decoder-only LLMs use an attention mask to limit the interaction
among tokens, there are no attention scores between the minority label
and some features if the minority label is shuffled to the middle
of the sequence. We address this problem by \textit{permuting only
the features while fixing the minority label at the beginning}. Finally,
after generating minority samples $\hat{x}$, other LLM-based methods
often require a verification process to predict again the labels of
synthetic minority samples using either an external classifier \cite{zhang2023generative}
or the fine-tuned LLM \cite{nguyen2024tabular,yang2024language}.
However, this step is often unreliable due to the poor classifier/LLM
trained/fine-tuned with the imbalanced dataset. We fix it by \textit{fine-tuning
the LLM with only minority samples and interpolated samples}.

To summarize, our main contributions are as follows:
\begin{enumerate}
\item We propose \textbf{ImbLLM}, a novel LLM-based oversampling method
that introduces three key improvements---sampling, permutation, and
fine-tuning---to generate diverse and robust minority samples.
\item We conduct extensive experiments on 10 tabular datasets against eight
SOTA baselines. ImbLLM achieves the best performance in 5 cases and
ranks second in 3 cases.
\item Beyond improving imbalanced classification, we demonstrate that the
generated samples are superior in both quality and diversity.
\item We provide a theoretical analysis through an entropy-based perspective,
formally showing that our generation process promotes diversity.
\end{enumerate}

\section{Background\label{sec:Background}}

\subsection{Oversampling for imbalanced datasets\label{subsec:Oversampling-for-imbalanced}}

Given an \textit{imbalanced} dataset ${\cal D}_{train}=\{x_{i},y_{i}\}_{i=1}^{N}$,
each row is a pair of a sample $x_{i}$ with $M$ features $\{X_{1},...,X_{M}\}$
and a label $y_{i}$ (a value of the target variable $Y$). To simplify
the problem, we consider $Y$ as a binary variable i.e. $y_{i}\in\{0,1\}$
and ${\cal D}_{train}$ contains a set of \textit{majority} samples
${\cal D}_{major}$ and a set of \textit{minority} samples ${\cal D}_{minor}$,
where $\mid{\cal D}_{minor}\mid\ll\mid{\cal D}_{major}\mid$. In case
the dataset has multiple classes, we define the majority class as
the class with the highest number of samples and the minority class
as the class with the lowest number of samples.

Our goal is to learn a data synthesizer $G$ from ${\cal D}_{train}$,
then use $G$ to generate \textit{synthetic} minority samples $\hat{{\cal D}}_{minor}$
such that $\mid\hat{{\cal D}}_{minor}\mid=\mid{\cal D}_{major}\mid$.
Following other works \cite{chawla2002smote,zhang2023generative,yang2024language},
we use $\hat{{\cal D}}_{minor}$ to create the \textit{rebalanced}
dataset $\hat{{\cal D}}_{train}={\cal D}_{major}\cup\hat{{\cal D}}_{minor}$.
Finally, we evaluate the quality of $\hat{{\cal D}}_{minor}$ by measuring
the F1 and AUC scores of a ML classifier trained on $\hat{{\cal D}}_{train}$
and tested on a held-out dataset ${\cal D}_{test}$ (shown in Figure
\ref{fig:Training-and-evaluation}). A better score means a better
$\hat{{\cal D}}_{minor}$.

\subsection{LLM-based methods for oversampling\label{subsec:LLM-based-methods-for-oversampling}}

Existing LLM-based methods for oversampling \cite{Borisov2023,zhang2023generative,nguyen2024tabular,yang2024language}
have three steps: (1) fine-tuning a pre-trained LLM, (2) generating
synthetic minority samples, and (3) verifying generated minority samples.

\subsubsection{Fine-tuning}

First, these methods convert each sample $x_{i}$ and its label $y_{i}$
into a \textit{sentence}. Although there are several ways to construct
a sentence \cite{hegselmann2023tabllm}, they often transform the
$i$\textsuperscript{th} row $r_{i}=[X_{1}=v_{i,1},...,X_{M}=v_{i,M},Y=y_{i}]$
into a corresponding sentence $s_{i}=\text{"}X_{1}\text{ is }v_{i,1},...,X_{M}\text{ is }v_{i,M},Y\text{ is }y_{i}\text{"}$,
where $\{X_{1},...,X_{M}\}$ are feature names, $v_{i,j}$ is the
value of the $j$\textsuperscript{th} feature of the $i$\textsuperscript{th}
row, $Y$ is the target variable and $y_{i}$ is its value. For example,
the row $[Edu=Bachelor,Job=Sales,WH=35.5,Income<200K]$ is converted
into ``\textit{Edu is Bachelor, Job is Sales, WH is 35.5, Income
is <200K}''. Note that ``\textit{WH}'' stands for ``Working hours
per week''.

Second, as there is no spacial locality between the columns in the
imbalanced dataset, they permute both the features $X$ and the target
variable $Y$. We call this strategy \textit{permute\_xy}. Given a
sentence $s_{i}=\text{"}X_{1}\text{ is }v_{i,1},...,X_{M}\text{ is }v_{i,M},Y\text{ is }y_{i}\text{"}$,
let $a_{i,j}=\text{"}X_{j}\text{ is }v_{i,j}\text{"}$ with $j\in\{1,...,M\}$
and $a_{i,M+1}=\text{"}Y\text{ is }y_{i}\text{"}$, then the sentence
can be presented as $s_{i}=\text{"}a_{i,1},...,a_{i,M},a_{i,M+1}\text{"}$.
They apply a \textit{permutation function} $P$ to randomly shuffle
the order of $X$ and $Y$, which results in $s_{i}=\text{"}a_{i,k_{1}},...,a_{i,k_{M}},a_{i,k_{M+1}}\text{"}$,
where $[k_{1},...,k_{M},k_{M+1}]=P([1,...,M,M+1])$. For example,
the sentence ``\textit{Edu is Bachelor, Job is Sales, WH is 35.5,
Income is <200K}'' is permuted to ``\textit{Job is Sales, Edu is
Bachelor, Income is <200K, WH is 35.5}''.

Finally, they fine-tune a pre-trained LLM following an auto-regressive
manner. Given the imbalanced dataset in the text format ${\cal S}=\{s_{i}\}_{i=1}^{N}$,
for each sentence $s_{i}\in{\cal S}$, they tokenize it into a sequence
of tokens $(c_{1},...,c_{l})=\text{tokenize}(s_{i})$. They factorize
the probability of a sentence $s_{i}$ into a product of output probabilities
conditioned on previously observed tokens:
\begin{equation}
p(s_{i})=p(c_{1},...,c_{l})=\prod_{k=1}^{l}p(c_{k}\mid c_{1},...,c_{k-1})\label{eq:fine-tune}
\end{equation}

They train the model using maximum likelihood estimation (MLE), which
maximizes the probability $\prod_{s_{i}\in{\cal S}}p(s_{i})$ of the
entire training dataset ${\cal S}$.

\subsubsection{Sampling $\hat{x}$}

First, they use the minority label ``$Y$ is $y_{minor}$'' as a
\textit{condition} to query the fine-tuned LLM $Q$ e.g. ``\textit{Income
is $\geq200$K}''. This condition is converted into a sequence of
tokens $(c_{1},...,c_{k-1})$. Second, they sample the next token
$c_{k}$ from a conditional probability distribution defined by a
\textit{softmax} function:
\begin{equation}
p(c_{k}\mid c_{1},...,c_{k})=\frac{e^{(z_{c_{k}}/T)}}{\sum_{c'\in{\cal C}}e^{(z_{c'}/T)}},\label{eq:sampling}
\end{equation}
where $e()$ is an exponential function, $z=Q(c_{1},...,c_{k-1})$
are the logits over all possible tokens, $T>0$ is a temperature,
and ${\cal C}$ is the vocabulary of tokens.

This step is repeated until all the features are sampled to generate
a synthetic minority sample $\hat{x}$. We call this strategy \textit{condition\_y}.

\subsubsection{Verifying $\hat{x}$}

For a standard LLM-based method like Great \cite{Borisov2023}, the
process to generate $\hat{x}$ is stopped at step (2). Some methods
employ the third step to verify generated minority samples. Given
a synthetic minority sample $\hat{x}_{i}=[X_{1}=v_{i,1},...,X_{M}=v_{i,M}]$,
TapTap \cite{zhang2023generative} uses a classifier trained with
${\cal D}_{train}$ to predict the label $\hat{y}_{i}$ for $\hat{x}_{i}$
again. Meanwhile, Pred-LLM \cite{nguyen2024tabular} and LITO \cite{yang2024language}
construct a prompt based on $\hat{x}_{i}$ as ``$X_{1}\text{ is }v_{i,1},...,X_{M}\text{ is }v_{i,M}$''
and use it as a condition to query the fine-tuned LLM $Q$ for the
label $\hat{y}_{i}$. If $\hat{y}_{i}$ is predicted as a majority
label, they discard $\hat{x}_{i}$.

\section{Framework\label{sec:Framework}}

\subsection{The proposed method}

We propose an LLM-based method (called \textbf{ImbLLM}) to generate
minority samples that mimic the real minority samples. ImbLLM improves
the diversity of synthetic minority samples through the three following
proposals.

\subsubsection{Sampling $\hat{x}$ conditioned on both minority label and feature\label{subsec:Sampling-yx}}

As described in the section Background \ref{subsec:LLM-based-methods-for-oversampling},
generating synthetic minority samples $\hat{x}$ is straightforward
for LLM-based methods. By forming the minority label as initial tokens
to query the fine-tuned LLM $Q$, we can generate more minority samples,
which oversample the minority class and rebalance the dataset. However,
there are two problems with this approach we need to solve.

\textbf{First problem:} \textit{The synthetic minority samples may
not be diverse}. Assume that we want to synthesize 1,000 minority
samples, then we need to create 1,000 prompts where each of them is
the same sentence ``$Y\text{ is }y_{minor}$'', where $Y$ is the
target variable and $y_{minor}$ is the minority label e.g. ``\textit{Income
is $\geq200$K}''. We use Equation (\ref{eq:sampling}) to generate
the next token. We call this strategy \textit{condition\_y}. As it
is a softmax function, it may always return the token with the highest
probability. In other words, given 1,000 identical prompts, we may
receive very similar 1,000 synthetic minority samples.

To solve this problem, we propose a simple method that constructs
diverse prompts using both minority label and features. For each prompt
``$Y\text{ is }y_{minor}$'', we first uniformly sample a feature
$X_{i}$ from the list of features $\{X_{1},...,X_{M}\}$, then sample
a value from the distribution of $X_{i}$ (i.e. $v_{i}\sim p(X_{i})$),
finally concatenate ``$X_{i}\text{ is }v_{i}$'' with ``$Y\text{ is }y_{minor}$''.
Our prompt has a new form of ``$Y\text{ is }y_{minor},X_{i}\text{ is }v_{i}$''
e.g. ``\textit{Income is $\geq200$K,} \textit{Edu is Bachelor}''.
Our prompts are much more diverse as they combine the minority label
with different features and values. We call our strategy \textit{condition\_yx}.

\textbf{Second problem:} \textit{Using the minority label as initial
tokens may not generate diverse samples due to the permutation strategy
permute\_xy}. Recall that \textit{permute\_xy} is used in existing
LLM-based methods to shuffle both the features $X$ and the target
variable $Y$ (see the section Background \ref{subsec:LLM-based-methods-for-oversampling}).
As these methods use decoder-only LLMs, they use \textit{attention
matrices} to determine the influence among tokens. Formally, given
a sequence of tokens $(c_{1},...,c_{l})$, let $\mathbf{E}\in\mathbb{R}^{l\times d}$
be the embedding matrix of the tokens at any given transformer layer.
The self-attention mechanism computes the query and key matrices $\mathbf{Q},\mathbf{K}\in\mathbb{R}^{l\times d_{k}}$
using a learned linear transformation:
\begin{equation}
\mathbf{Q}=\mathbf{E}\mathbf{W}_{Q},\mathbf{K}=\mathbf{E\mathbf{W}}_{K},\label{eq:query-key-value}
\end{equation}
where $\mathbf{W}_{Q},\mathbf{W}_{K}$ are weight matrices learned
by the transformer model, and $d,d_{k}$ are dimensions of embedding
and query/key.

The attention matrix is computed as:
\begin{equation}
\mathcal{A}=\text{softmax}(\frac{{\cal M}_{\{i\leq j\}}\mathbf{Q}\mathbf{K}^{\text{T}}}{\sqrt{d_{k}}})\label{eq:attention-matrix}
\end{equation}

In Equation (\ref{eq:attention-matrix}), decoder-only LLMs use an
attention mask ${\cal M}_{\{i\leq j\}}$ to limit token interactions,
which only allows each token is influenced by its preceding tokens.
Namely, given a current token $i$, ${\cal M}_{\{i\leq j\}}=\mathbb{I}_{i\leq j}$
will set the attention score to $0$ if the token $j$ is on the left
of $i$.

As existing LLM-based methods permute both $X$ and $Y$, the minority
label $y_{minor}$ can be shuffled to the middle/end of the sentence.
Due to the attention mask, there is no attention score between $y_{minor}$
and some features before it. When we use $y_{minor}$ as a condition
to sample other features, LLMs have only few choices to generate the
next token, leading to less diverse synthetic samples.

\subsubsection{Permuting $X$ but fixing $Y$\label{subsec:Permuting-xy}}

To address the above problem, we propose the permutation strategy
\textit{fix\_y}, which only permutes $X$ while fixing $Y$ at the
beginning of the sentence. Formally, given a sentence $s_{i}=\text{"}a_{i,1},...,a_{i,M},a_{i,M+1}\text{"}$,
we apply the permutation function $P$ to $M$ features and move $Y$
to the beginning, which results in a permuted sentence $s_{i}=\text{"}a_{i,M+1},a_{i,k_{1}},...,a_{i,k_{M}}\text{"}$,
where $[k_{1},...,k_{M}]=P([1,...,M])$. For example, the sentence
``\textit{Edu is Bachelor, Job is Doctor, WH is 40.2, Income is $\geq200$K}''
is permuted to ``\textit{Income is $\geq200$K, Edu is Bachelor,
WH is 40.2, Job is Doctor}''. Note that ``\textit{Income is $\geq200$K}''
is at the beginning of the permuted sentence.

\begin{figure*}
\begin{centering}
\subfloat[\textit{permute\_xy}]{\begin{centering}
\includegraphics[scale=0.4]{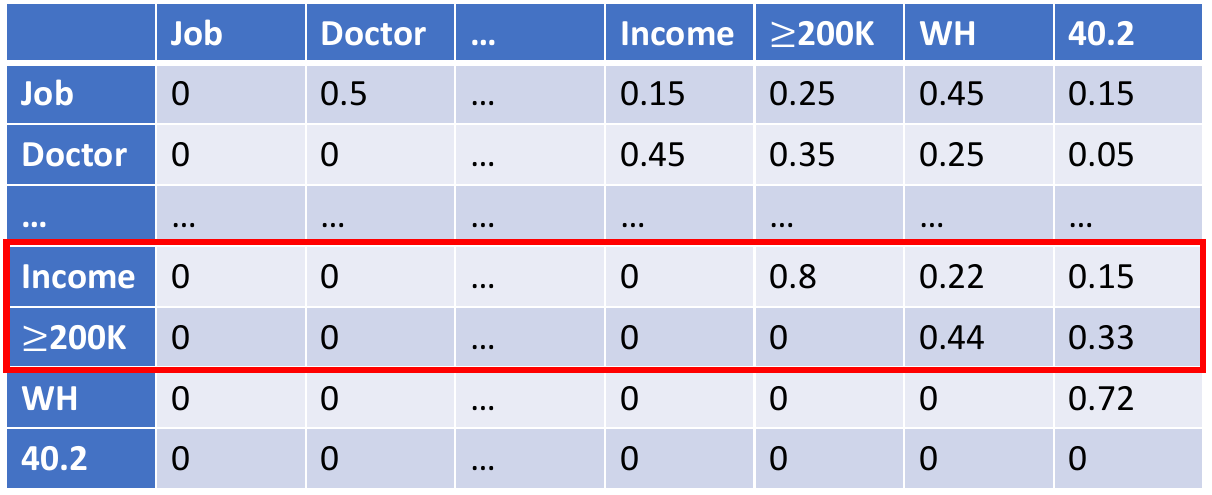}
\par\end{centering}
}\subfloat[\textit{fix\_y}]{\begin{centering}
\includegraphics[scale=0.4]{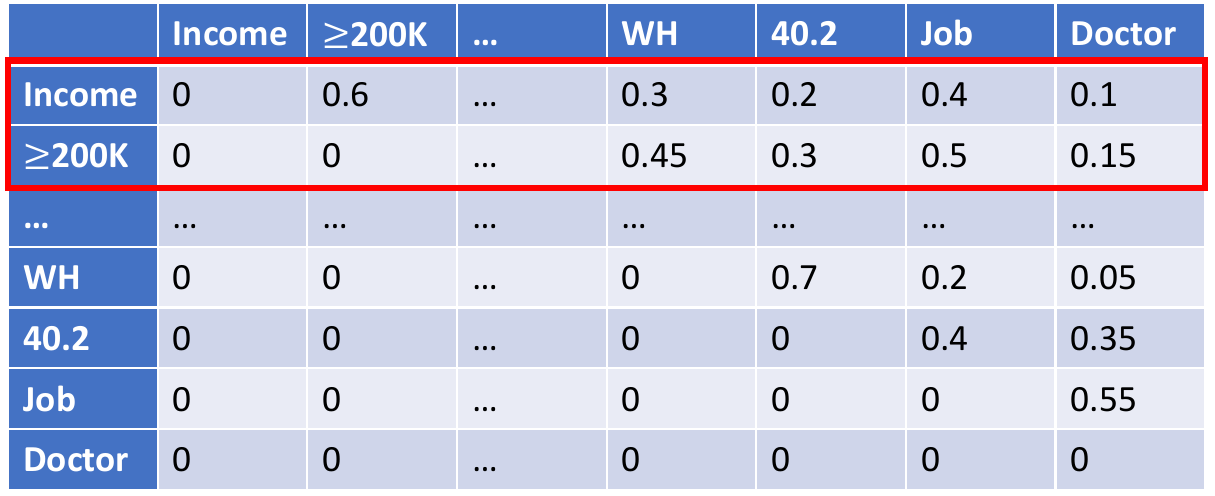}
\par\end{centering}
}
\par\end{centering}
\caption{\label{fig:Attention-matrix}Attention matrix. Existing LLM-based
methods use \textit{permute\_xy} to shuffle both $X$ and $Y$, leading
to no attention scores between $y_{minor}$ and some features (a).
Our method uses \textit{fix\_y} to permute only $X$ while fixing
$Y$ at the beginning, resulting in attention scores between $y_{minor}$
and all features (b).}
\end{figure*}

We illustrate the attention matrices derived by \textit{permute\_xy}
(other methods) and \textit{fix\_y} (our method) in Figure \ref{fig:Attention-matrix}.
Figure \ref{fig:Attention-matrix}(a) shows that using \textit{permute\_xy},
the minority label ``\textit{$\geq200$K}'' has only attention score
with the feature ``\textit{WH}'', and the LLM is likely to predict
``\textit{WH is 40.2}'' after the condition ``\textit{Income is
$\geq200$K}''. In contrast, Figure \ref{fig:Attention-matrix}(b)
shows that our \textit{fix\_y} helps the minority label ``\textit{$\geq200$K}''
to retrieve attention scores with all features, and the LLM can predict
any feature after the condition, resulting in more diverse synthetic
samples.

\subsubsection{Fine-tunning LLMs with minority samples and their interpolation\label{subsec:Fine-tunning-LLMs}}

As existing LLM-based methods use both majority and minority samples
to fine-tune the LLM, the synthetic minority samples $\hat{x}$ may
be biased to the majority label although they are already conditioned
on the minority label \cite{yang2024language}. They address this
problem by using either an external classifier \cite{zhang2023generative}
or the fine-tuned LLM itself \cite{nguyen2024tabular,yang2024language}
to verify $\hat{x}$. But this strategy raises another problem. Since
the classifier/LLM is trained/fine-tuned with the imbalanced dataset
${\cal D}_{train}$, we expect that their performance is not optimal,
leading to an unreliable verification step. As we show in the experiments,
this verification does not work on nearly half of the benchmark datasets.

We propose a simple trick to make sure that the generated samples
are indeed minority samples without involving any verification. Instead
of using both majority and minority samples, we use only minority
samples to fine-tune the LLM. This helps our LLM to focus on generating
minority samples while it is not influenced by (or biased to) majority
samples. However, it has a weakness. The number of possible values
of a continuous variable is often much larger than that of a categorical
variable. For example, the categorical variable ``\textit{Edu}''
has only few values \{``School'', ``Bachelor'', ``Master'',
``PhD''\} whereas the continuous variable ``\textit{WH}'' can
have any continuous value in $[0,168]$. Thus, if we use ${\cal D}_{minor}$
(a small dataset), although we have only few samples, their categorical
values are still good enough to cover the categorical domains. However,
their continuous values are too limited to capture the characteristics
of the continuous domains.

\textbf{Interpolation.} We address this issue with an interpolation
step. First, we randomly sample a minority sample $x_{i}\in{\cal D}_{minor}$.
Let $x_{i}^{con}$ and $x_{i}^{cat}$ be continuous and categorical
values of $x_{i}$. Then, we interpolate $x_{i}$ using:
\begin{equation}
x_{i}'=x_{i}^{con}+\epsilon(x_{j}^{con}-x_{i}^{con}),\label{eq:interpolate}
\end{equation}
where $x_{j}\in{\cal D}_{minor}$ is another random minority sample
and $\epsilon\in[0,1]$ is a random number. Here, we only interpolate
continuous variables of $x_{i}$. After the interpolation process,
we obtain a set of \textit{interpolated samples} ${\cal D}_{inter}=\{x'_{i},y'_{i}\}_{i=1}^{\mid{\cal D}_{major}\mid}$.
Note that a minority sample $x_{i}\in{\cal D}_{minor}$ has $M$ features
but an interpolated sample $x_{i}'\in{\cal D}_{inter}$ has only $M_{con}$
features where $M_{con}$ and $M_{cat}$ are the numbers of continuous
and categorical variables and $M_{con}+M_{cat}=M$. We generate the
set of interpolated samples such that $\mid{\cal D}_{inter}\mid=\mid{\cal D}_{major}\mid$.
Finally, we fine-tune our LLM with ${\cal D}_{train}={\cal D}_{minor}\cup{\cal D}_{inter}$.

\begin{figure*}[t]
\begin{centering}
\includegraphics[scale=0.42]{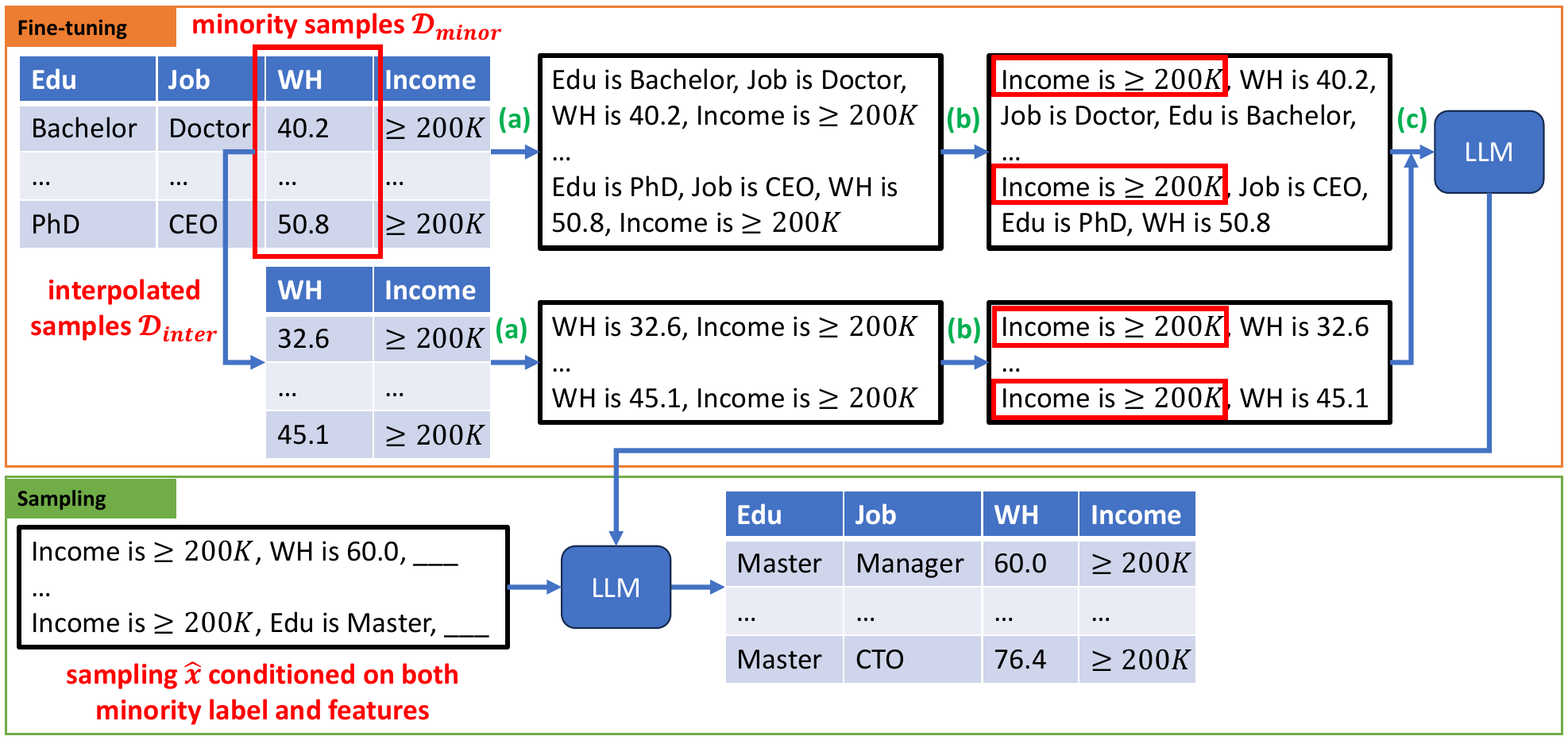}
\par\end{centering}
\caption{\label{fig:Overview-of-ImbLLM}Overview of our ImbLLM. \textbf{First
step:} it converts each row to a sentence (a), permutes $X$ and fixes
$Y$ at the beginning (b), and fine-tunes an LLM with minority samples
${\cal D}_{minor}$ and interpolated samples ${\cal D}_{inter}$ (c).
\textbf{Second step:} it constructs prompts based on both minority
label and features to query the fine-tuned LLM to generate minority
samples $\hat{x}$.}
\end{figure*}

The workflow of our method ImbLLM is illustrated in Figure \ref{fig:Overview-of-ImbLLM}.

\section{Experiments\label{sec:Experiments}}

We conduct extensive experiments to show that our method is better
than other methods under different qualitative and quantitative metrics.

\subsection{Experiment settings}

\subsubsection{Datasets}

We evaluate our method on 10 real-world tabular datasets. They are
commonly used in classification tasks \cite{Xu2019,Borisov2023,zhang2023generative}.
The details of these datasets are provided in Table \ref{tab:Characteristics-datasets}.

\begin{table}[H]
\caption{\label{tab:Characteristics-datasets}Dataset characteristics. $N$,
$M$, \textit{\#Con},\textit{ \#Cat}, and $K$ indicate the numbers
of samples, features, continuous variables, categorical variables,
and classes.}

\centering{}%
\begin{tabular}{|l|r|r|r|r|r|l|}
\hline 
\rowcolor{header_color}Dataset & $N$ & $M$ & \textit{\#Con} & \textit{\#Cat} & $K$ & Source\tabularnewline
\hline 
\hline 
\textit{adult} & 30,162 & 11 & 5 & 6 & 2 & UCI\tabularnewline
\hline 
\rowcolor{even_color}\textit{german} & 1,000 & 20 & 6 & 14 & 2 & UCI\tabularnewline
\hline 
\textit{fuel} & 639 & 8 & 3 & 5 & 2 & Kaggle\tabularnewline
\hline 
\rowcolor{even_color}\textit{insurance} & 1,338 & 6 & 3 & 3 & 2 & Kaggle\tabularnewline
\hline 
\textit{gem\_price} & 26,270 & 9 & 6 & 3 & 2 & Kaggle\tabularnewline
\hline 
\rowcolor{even_color}\textit{housing} & 20,433 & 9 & 8 & 1 & 2 & Kaggle\tabularnewline
\hline 
\textit{bank} & 4,521 & 14 & 5 & 9 & 2 & \cite{Nguyen2021}\tabularnewline
\hline 
\rowcolor{even_color}\textit{car} & 1,728 & 6 & 2 & 6 & 4 & UCI\tabularnewline
\hline 
\textit{credit\_card} & 10,127 & 19 & 14 & 5 & 2 & UCI\tabularnewline
\hline 
\rowcolor{even_color}\textit{sick} & 3,560 & 27 & 6 & 21 & 2 & OpenML\tabularnewline
\hline 
\end{tabular}
\end{table}

\subsubsection{Evaluation metric\label{subsec:Evaluation-metric}}

To evaluate the performance of oversampling methods, we use the synthetic
minority samples to rebalance the imbalanced dataset similar to other
works \cite{Xu2019,Borisov2023,zhang2023generative}. For each dataset,
we randomly split it into 80\% for the training set ${\cal D}_{train}$
and 20\% for the test set ${\cal D}_{test}$. To construct an imbalanced
dataset, we reduce the number of minority samples like other works\cite{zhang2023generative,yang2024language},
where we only use 20\% minority samples. Formally, the \textit{imbalanced}
training set ${\cal D}_{train}={\cal D}_{major}\cup{\cal D}_{minor}$,
where $\mid{\cal D}_{minor}\mid\ll\mid{\cal D}_{major}\mid$ and $\mid{\cal D}{}_{minor}\mid=q\times\mid{\cal D}_{minor}^{*}\mid$
with $q\in(0,1]$ being the \emph{imbalance-ratio}\textit{\emph{ and
${\cal D}_{minor}^{*}$ being the }}\emph{original}\textit{\emph{
set of minority samples (before reduced). We }}set $q=0.2$ for all
experiments.

We then train oversampling methods with the \emph{imbalanced} dataset
${\cal D}{}_{train}$ to generate the set of synthetic minority samples
$\hat{{\cal D}}{}_{minor}$ such that $\mid\hat{{\cal D}}{}_{minor}\mid=\mid{\cal D}_{major}\mid$.
Finally, we train the ML classifier XGBoost on the \emph{rebalanced}
dataset $\hat{{\cal D}}{}_{train}={\cal D}_{major}\cup\hat{{\cal D}}{}_{minor}$
and compute its F1-score and AUC on ${\cal D}_{test}$. We repeat
each method \textit{three times with random seeds} and report the
average score along with its standard deviation.

We also compute other metrics between the \textit{original} minority
samples ${\cal D}_{minor}^{*}$ and the \emph{synthetic} minority
samples $\hat{{\cal D}}{}_{minor}$ to measure the \emph{quality}
and \textit{diversity} of $\hat{{\cal D}}{}_{minor}$. Note that ${\cal D}_{minor}^{*}$
is \textbf{\emph{unseen}} for all oversampling methods. It is just
used for the evaluation purpose.

(1) \emph{Close probability} \cite{qian2023synthcity}: We compute
the probability of \emph{close values} between ${\cal D}_{minor}^{*}$
and $\hat{{\cal D}}{}_{minor}$. For each real minority sample $x\in{\cal D}_{minor}^{*}$,
we compute its distance to the closest synthetic minority sample $\hat{x}\in\hat{{\cal D}}{}_{minor}$,
and normalize the distance $d(x,\hat{x})$ to $[0,1]$. Given a threshold
$\alpha=0.2$, we define $x$ has a close value if $d(x,\hat{x})\leq\alpha$.
We compute the close probability as $Close=\frac{1}{N}\sum_{i=1}^{N}\mathbb{I}_{d(x_{i},\hat{x}_{i})\leq\alpha}$.
This score measures the \emph{quality} of $\hat{{\cal D}}{}_{minor}$,
where $0$ means synthetic samples are very dissimilar to real samples
whereas $1$ means all synthetic samples are similar to some real
samples. \textit{A higher score is better}.

(2) \textit{Coverage} \cite{naeem2020reliable}: We compute the coverage
score between ${\cal D}_{minor}^{*}$ and $\hat{{\cal D}}{}_{minor}$,
which measures the fraction of real samples whose neighborhoods contain
at least one synthetic sample. $Cov=\frac{1}{N}\sum_{i=1}^{N}\mathbb{I}_{\exists j\text{ s.t. }\hat{x}_{j}\in B(x_{i},d_{k}(x_{i}))}$,
where $B(x_{i},d_{k}(x_{i}))$ is the sphere around $x_{i}$ and the
radius $d_{k}(x_{i})$ is the Euclidean distance from $x_{i}$ to
its $k$-th nearest neighbor (we set $k=2$). It measures the \textit{diversity}
of $\hat{{\cal D}}{}_{minor}$ as it shows how the synthetic minority
samples resemble the variability of the real minority samples. \textit{A
higher score is better}.

\subsubsection{Baselines}

We compare with eight oversampling baselines, including traditional
methods (\emph{AdaSyn} \cite{he2008adasyn}, \emph{SMOTE} \cite{chawla2002smote},
and \emph{SMOTE-NC} \cite{chawla2002smote}), generative-based methods
(\textit{CTGAN} \cite{Xu2019} and \textit{TVAE} \cite{Xu2019}),
and LLM-based methods (\textit{Great} \cite{Borisov2023}, \textit{TapTap}
\cite{zhang2023generative}, and \emph{Pred-LLM} \cite{nguyen2024tabular}).
For reproducibility, we use the published source codes of AdaSyn,
SMOTE, and SMOTE-NC\footnote{https://imbalanced-learn.org/stable/},
TVAE and CTGAN\footnote{https://github.com/sdv-dev/CTGAN}, Great\footnote{https://github.com/kathrinse/be\_great},
TapTap\footnote{https://github.com/ZhangTP1996/TapTap}, and Pred-LLM\footnote{https://github.com/nphdang/Pred-LLM}.
We do not compare with the LLM-based method LITO \cite{yang2024language}
as its source code is not available.

The \textit{Original} method means the classifier trained with the
\emph{original} majority and minority sets ${\cal D}_{major}$ and
${\cal D}_{minor}^{*}$. As ${\cal D}_{minor}^{*}$ is large and just
slightly imbalanced from ${\cal D}_{major}$, we can expect the Original
method achieves a good result, which can be served as the reference.
The \emph{Imbalance} method means the classifier trained directly
with the \emph{imbalanced}\textit{ dataset} ${\cal D}{}_{train}$.
Other methods mean the classifiers trained with the \emph{rebalanced}\textit{
dataset} $\hat{{\cal D}}{}_{train}$ after performing the oversampling
step. We use XGBoost for the classifier since it is one of the most
popular predictive models for tabular data \cite{shwartz2022tabular}.
Following \cite{Borisov2023,zhang2023generative,nguyen2024tabular},
we use the distilled version of ChatGPT-2 for our LLM, train it with
\textit{batch\_size=32}, \textit{\#epochs=50}, and set $T=0.7$ for
Equation (\ref{eq:sampling}).

\subsection{Results and discussions}

\subsubsection{Imbalanced classification task}

Table \ref{tab:Main-experiments} reports F1-score of each oversampling
method on 10 benchmark datasets. \textit{We recall that the procedure
to compute the score is described in Figure \ref{fig:Training-and-evaluation}}.
Our method ImbLLM performs better than other methods. Among 10 datasets,
ImbLLM is the best performing on five datasets and the second-best
on another three datasets. The average improvement of ImbLLM over
CTGAN (the runner-up method) is around $3\%$ and Imbalance (the standard
classifier without oversampling) is around $8\%$.

\begin{table*}
\caption{\label{tab:Main-experiments}F1-score $\pm$ (std) of each oversampling
method on 10 datasets. \textbf{Bold} and \uline{underline} indicate
the best and second-best methods.}

\centering{}%
\begin{tabular}{|l|ccccccccccc|}
\hline 
\rowcolor{header_color}F1-score & Original & Imbalance & AdaSyn & SMOTE & SMOTE-NC & CTGAN & TVAE & Great & TapTap & Pred-LLM & ImbLLM\tabularnewline
\hline 
\hline 
\textit{\emph{adult}} & 0.6761 & 0.4685 & 0.3987 & 0.3987 & 0.5586 & 0.5223 & 0.4985 & \textbf{0.6592} & 0.4705 & 0.5071 & \uline{0.6543}\tabularnewline
 & (0.056) & (0.030) & (0.000) & (0.000) & (0.056) & (0.013) & (0.041) & (0.040) & (0.024) & (0.059) & (0.044)\tabularnewline
\hline 
\rowcolor{even_color}\textit{\emph{german}} & 0.5299 & 0.2032 & \textbf{0.4601} & \uline{0.4595} & 0.2706 & 0.4064 & 0.2760 & 0.2059 & 0.2069 & 0.2363 & 0.4419\tabularnewline
\rowcolor{even_color} & (0.080) & (0.076) & (0.002) & (0.003) & (0.127) & (0.039) & (0.124) & (0.148) & (0.060) & (0.034) & (0.053)\tabularnewline
\hline 
\textit{\emph{fuel}} & 0.9645 & 0.9664 & 0.9458 & 0.9406 & 0.9662 & 0.9658 & 0.9642 & 0.9272 & \uline{0.9751} & 0.9662 & \textbf{0.9820}\tabularnewline
 & (0.026) & (0.018) & (0.038) & (0.041) & (0.009) & (0.009) & (0.019) & (0.017) & (0.011) & (0.010) & (0.017)\tabularnewline
\hline 
\rowcolor{even_color}\textit{\emph{insurance}} & 0.9153 & 0.9220 & 0.3933 & 0.4263 & \textbf{0.9253} & 0.8858 & 0.9193 & 0.7690 & 0.9209 & 0.9182 & \uline{0.9250}\tabularnewline
\rowcolor{even_color} & (0.005) & (0.006) & (0.054) & (0.027) & (0.005) & (0.034) & (0.009) & (0.016) & (0.012) & (0.010) & (0.019)\tabularnewline
\hline 
\textit{\emph{gem\_price}} & 0.9799 & 0.9707 & 0.2818 & 0.5931 & \uline{0.9739} & 0.9705 & 0.9711 & 0.9698 & 0.9709 & 0.9733 & \textbf{0.9755}\tabularnewline
 & (0.002) & (0.002) & (0.158) & (0.147) & (0.001) & (0.002) & (0.001) & (0.003) & (0.002) & (0.002) & (0.001)\tabularnewline
\hline 
\rowcolor{even_color}\textit{\emph{housing}} & 0.9111 & 0.8738 & 0.4421 & 0.4450 & \uline{0.8801} & 0.8690 & 0.8707 & 0.8721 & 0.8756 & \textbf{0.8826} & 0.8750\tabularnewline
\rowcolor{even_color} & (0.002) & (0.006) & (0.005) & (0.005) & (0.011) & (0.011) & (0.016) & (0.009) & (0.005) & (0.009) & (0.011)\tabularnewline
\hline 
\textit{\emph{bank}} & 0.3657 & 0.1118 & 0.2063 & 0.2064 & 0.2322 & \uline{0.3162} & 0.2134 & 0.2183 & 0.0831 & 0.1352 & \textbf{0.3441}\tabularnewline
 & (0.035) & (0.022) & (0.000) & (0.000) & (0.077) & (0.071) & (0.047) & (0.025) & (0.029) & (0.102) & (0.028)\tabularnewline
\hline 
\rowcolor{even_color}\textit{\emph{car}} & 0.9923 & 0.9757 & 0.7695 & 0.7689 & 0.9684 & 0.9787 & 0.9734 & 0.5732 & 0.9716 & \textbf{0.9831} & \uline{0.9820}\tabularnewline
\rowcolor{even_color} & (0.004) & (0.011) & (0.006) & (0.007) & (0.010) & (0.005) & (0.007) & (0.034) & (0.014) & (0.005) & (0.004)\tabularnewline
\hline 
\textit{\emph{credit\_card}} & 0.9825 & 0.9671 & 0.0436 & 0.0701 & \uline{0.9725} & 0.9698 & 0.9671 & 0.9634 & 0.9634 & 0.9692 & \textbf{0.9758}\tabularnewline
 & (0.002) & (0.002) & (0.015) & (0.024) & (0.001) & (0.001) & (0.003) & (0.002) & (0.001) & (0.002) & (0.002)\tabularnewline
\hline 
\rowcolor{even_color}\textit{\emph{sick}} & 0.9208 & 0.7790 & 0.1826 & 0.2057 & 0.8270 & \uline{0.8394} & 0.8376 & 0.6752 & 0.7386 & 0.7740 & \textbf{0.8459}\tabularnewline
\rowcolor{even_color} & (0.032) & (0.012) & (0.053) & (0.061) & (0.009) & (0.040) & (0.037) & (0.044) & (0.026) & (0.030) & (0.004)\tabularnewline
\hline 
\rowcolor{childheader_color}Average & 0.8238 & 0.7238 & 0.4124 & 0.4514 & 0.7575 & \uline{0.7724} & 0.7491 & 0.6833 & 0.7177 & 0.7345 & \textbf{0.8002}\tabularnewline
\hline 
\end{tabular}
\end{table*}

\begin{figure*}[t]
\begin{centering}
\includegraphics[scale=0.48]{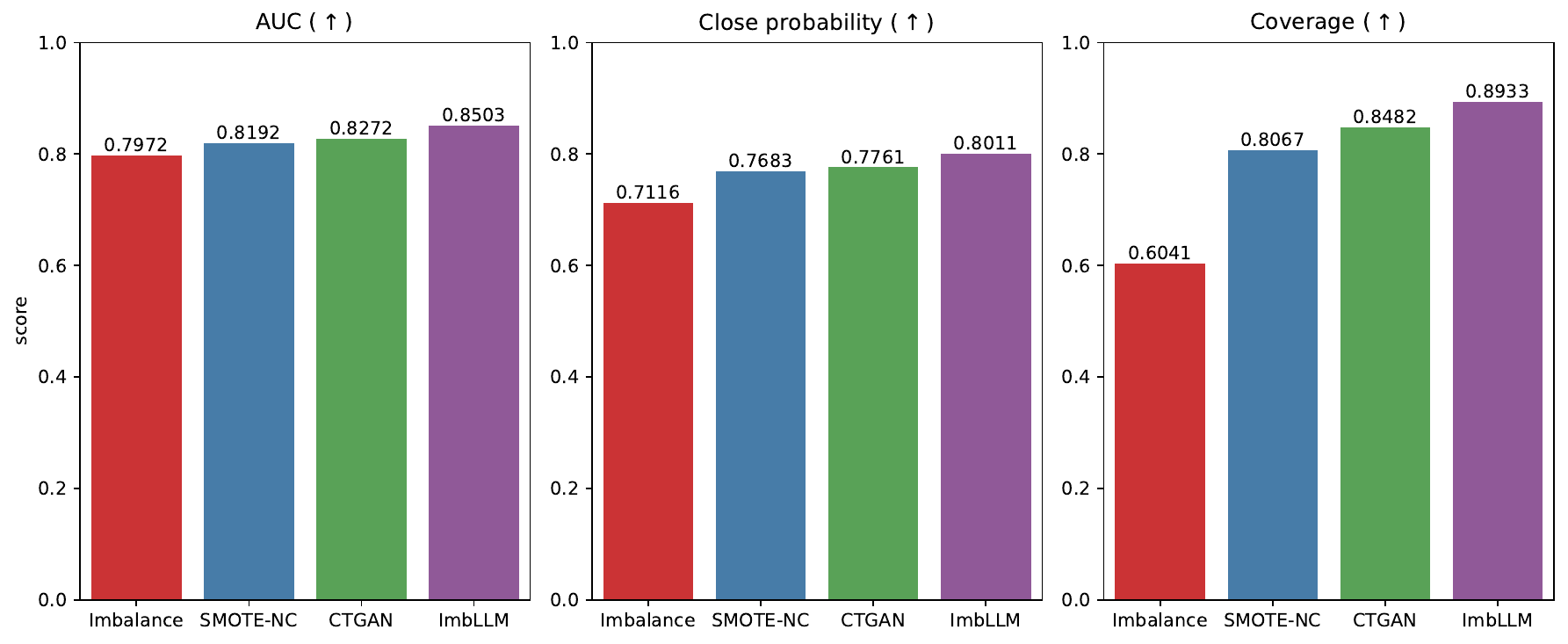}
\par\end{centering}
\caption{\label{fig:Quality-and-diversity}The average quality and diversity
scores. $\uparrow$ means \textquotedblleft\textit{higher is better}\textquotedblright .}
\end{figure*}

As we discussed earlier, traditional methods AdaSyn and SMOTE perform
poorly as they require a conversion of categorical variables to numerical
vectors, leading to information loss. This is a serious problem as
many experimental datasets have a large number of categorical variables.
SMOTE-NC (it is an improved version of SMOTE) addresses mixed types
much better and becomes the third-best method, following CTGAN. The
LLM-based methods are better than SMOTE and they are comparable with
TVAE. The results also show that TapTap and Pred-LLM with the verification
step via either an external classifier or the fine-tuned LLM significantly
outperform the standard LLM method Great. However, thanks to the diversity
of our synthetic minority samples, ImbLLM greatly improves over TapTap
and Pred-LLM by $9\%$ and $7\%$.

Most oversampling methods improve over Imbalance (the method trains
the classifier on the imbalanced dataset without rebalancing it).
However, their performance is still far away from the Original method.
Only our ImbLLM can reach to the performance level of Original (0.8002
vs. 0.8238) although ImbLLM only uses 20\% minority samples of Original.

The AUC scores of top-3 methods are reported in Figure \ref{fig:Quality-and-diversity},
which also shows ImbLLM is the best method.

\subsubsection{Quality and diversity evaluation}

As described in Section \ref{subsec:Evaluation-metric}, we compute
the Close probability and Coverage scores to measure the \textit{quality
and diversity of synthetic minority samples}. Figure \ref{fig:Quality-and-diversity}
presents the average scores of top-3 methods on all datasets. Our
method ImbLLM significantly outperforms other methods for all metrics.
The quality metric Close probability shows that our synthetic minority
samples are closer to the original minority samples than other synthetic
samples (0.8011 vs. 0.7761 of CTGAN). In terms of diversity, the Coverage
score shows that our synthetic minority samples are more diverse than
others (0.8933 vs. 0.8482 of CTGAN).

High scores in both quality and diversity metrics show our synthetic
minority samples can capture the \textit{real manifold of minority
samples}, which explains their benefits in imbalanced classification
tasks (see Table \ref{tab:Main-experiments} and ``AUC'' in Figure
\ref{fig:Quality-and-diversity}).

\subsection{Ablation studies}

We analyze our method under different configurations.

\subsubsection{Effect investigation of various modifications}

We have three novel contributions in \textit{sampling}, \textit{permutation},
and \textit{fine-tuning}. Table \ref{tab:Effectiveness-of-different}
reports the F1-score of each modification compared to that of the
LLM baseline Great. Recall that Great generates minority samples conditioned
on the minority label (denoted by \textit{condition\_y}), permutes
both the features $X$ and the target variable $Y$ (denoted by \textit{permute\_xy}),
and fine-tunes the LLM with both majority and minority samples (denoted
by \textit{major+minor}).

While Great only achieves 0.6833, our modifications show improvements.
By fixing the target variable $Y$ at the beginning of the sentence
(denoted by \textit{fix\_y}) or fine-tuning the LLM with only minority
samples and their interpolation (denoted by \textit{minor+interpolate}),
we achieve F1-score up to $\sim0.74$ (i.e. $6\%$ improvement). Combining
these two proposals with our diversity sampling strategy (denoted
by\textit{ condition\_yx}), we achieve the best result at 0.8002 (the
last row). This study suggests that each modification is useful, which
significantly improves Great.

\begin{table}[th]
\caption{\label{tab:Effectiveness-of-different}Effectiveness of different
modifications in ImbLLM. The F1-score is averaged over 10 datasets.
\textquotedblleft\checkmark\textquotedblright{} indicates ImbLLM
uses the same setting as Great.}

\centering{}%
\begin{tabular}{|l|l|l|l|r|}
\hline 
\rowcolor{header_color} & Sampling $\hat{x}$ & Permutation & Fine-tuning & \textbf{F1-score}\tabularnewline
\hline 
\hline 
Great & \textit{condition\_y} & \textit{permute\_xy} & \textit{major+minor} & 0.6833\tabularnewline
\hline 
\multirow{7}{*}{ImbLLM} & \textit{condition\_yx} & \checkmark & \checkmark & 0.6690\tabularnewline
\cline{2-5}
 & \checkmark & \textit{fix\_y} & \checkmark & 0.7408\tabularnewline
\cline{2-5}
 & \checkmark & \checkmark & \textit{minor+interpolate} & 0.7436\tabularnewline
\cline{2-5}
 & \textit{condition\_yx} & \textit{fix\_y} & \checkmark & 0.7181\tabularnewline
\cline{2-5}
 & \textit{condition\_yx} & \checkmark & \textit{minor+interpolate} & 0.7998\tabularnewline
\cline{2-5}
 & \checkmark & \textit{fix\_y} & \textit{minor+interpolate} & 0.7421\tabularnewline
\cline{2-5}
 & \textbf{\textit{condition\_yx}} & \textbf{\textit{fix\_y}} & \textbf{\textit{minor+interpolate}} & \textbf{0.8002}\tabularnewline
\hline 
\end{tabular}
\end{table}

\subsubsection{Effect investigation of interpolation}

As discussed earlier, we fine-tune the LLM using only minority samples
to avoid biased generated minority samples. This also helps us to
skip the verification step via either an external classifier (used
in TapTap) or the LLM itself (used in Pred-LLM). However, as the minority
set is small, the synthetic minority samples may not be generalizable
to cover \textit{unseen} minority samples, leading to poor performance
in imbalanced classification tasks. We address this problem with interpolated
samples (see Section \ref{subsec:Fine-tunning-LLMs}). We show the
benefit of the interpolation by comparing ImbLLM with its variation
\textit{ImbLLM-inter} (i.e. we keep all components in ImbLLM except
the interpolation step). The results are summarized in Table \ref{tab:ImbLLM-inter},
where ImbLLM is better than ImbLLM-inter on most datasets in terms
of both classification performance and quality. Interestingly, ImbLLM-inter
is still better than CTGAN (the best baseline) on all scores, which
proves our contributions are very effective and robust.

\begin{table*}
\caption{\label{tab:ImbLLM-inter}ImbLLM vs. ImbLLM-inter (i.e. ImbLLM without
interpolation).}

\centering{}%
\begin{tabular}{|l|c|c|c|c|c|c|}
\hline 
\rowcolor{header_color} & \multicolumn{2}{c|}{F1-score} & \multicolumn{2}{c|}{AUC} & \multicolumn{2}{c|}{Close probability}\tabularnewline
\hline 
\rowcolor{header_color} & ImbLLM-inter & ImbLLM & ImbLLM-inter & ImbLLM & ImbLLM-inter & ImbLLM\tabularnewline
\hline 
\hline 
\textit{\emph{adult}} & \textbf{0.6792} & 0.6543 & \textbf{0.7904} & 0.7662 & \textbf{0.9989} & 0.9982\tabularnewline
\hline 
\rowcolor{even_color}\textit{\emph{german}} & 0.3917 & \textbf{0.4419} & 0.6159 & \textbf{0.6357} & 0.9472 & \textbf{0.9528}\tabularnewline
\hline 
\textit{\emph{fuel}} & 0.9730 & \textbf{0.9820} & 0.9597 & \textbf{0.9716} & 0.3704 & \textbf{0.4237}\tabularnewline
\hline 
\rowcolor{even_color}\textit{\emph{insurance}} & 0.9118 & \textbf{0.9250} & 0.9025 & \textbf{0.9135} & \textbf{0.5569} & 0.4990\tabularnewline
\hline 
\textit{\emph{gem\_price}} & 0.9740 & \textbf{0.9755} & \textbf{0.9724} & 0.9723 & \textbf{0.9941} & 0.9925\tabularnewline
\hline 
\rowcolor{even_color}\textit{\emph{house}} & \textbf{0.8813} & 0.8750 & \textbf{0.8424} & 0.8388 & 0.9972 & \textbf{0.9977}\tabularnewline
\hline 
\textit{\emph{bank}} & 0.2544 & \textbf{0.3441} & 0.5757 & \textbf{0.6174} & \textbf{0.9928} & 0.9816\tabularnewline
\hline 
\rowcolor{even_color}\textit{\emph{car}} & 0.9768 & \textbf{0.9820} & 0.9626 & \textbf{0.9683} & 0.3526 & \textbf{0.3782}\tabularnewline
\hline 
\textit{\emph{credit\_card}} & 0.9724 & \textbf{0.9758} & 0.8732 & \textbf{0.8983} & 0.8902 & \textbf{0.9188}\tabularnewline
\hline 
\rowcolor{even_color}\textit{\emph{sick}} & 0.8426 & \textbf{0.8459} & 0.8637 & \textbf{0.9209} & 0.7704 & \textbf{0.8685}\tabularnewline
\hline 
\rowcolor{childheader_color}Average & 0.7857 & \textbf{0.8002} & 0.8359 & \textbf{0.8503} & 0.7871 & \textbf{0.8011}\tabularnewline
\hline 
\end{tabular}
\end{table*}

We also investigate the impact of the number of interpolated samples
on our method's performance. Recall that in Section \ref{subsec:Fine-tunning-LLMs},
we generate the set of interpolated samples ${\cal D}_{inter}=\{x'_{i},y'_{i}\}_{i=1}^{\mid{\cal D}_{major}\mid}$,
where the number of interpolated samples $\mid{\cal D}_{inter}\mid=\mid{\cal D}_{major}\mid$.
Here, we control the size of ${\cal D}_{inter}$ by employing an \textit{interpolation-ratio}
$r\in[0,1]$ such that $\mid{\cal D}_{inter}\mid=r\times\mid{\cal D}_{major}\mid$.

\begin{figure}
\begin{centering}
\includegraphics[scale=0.5]{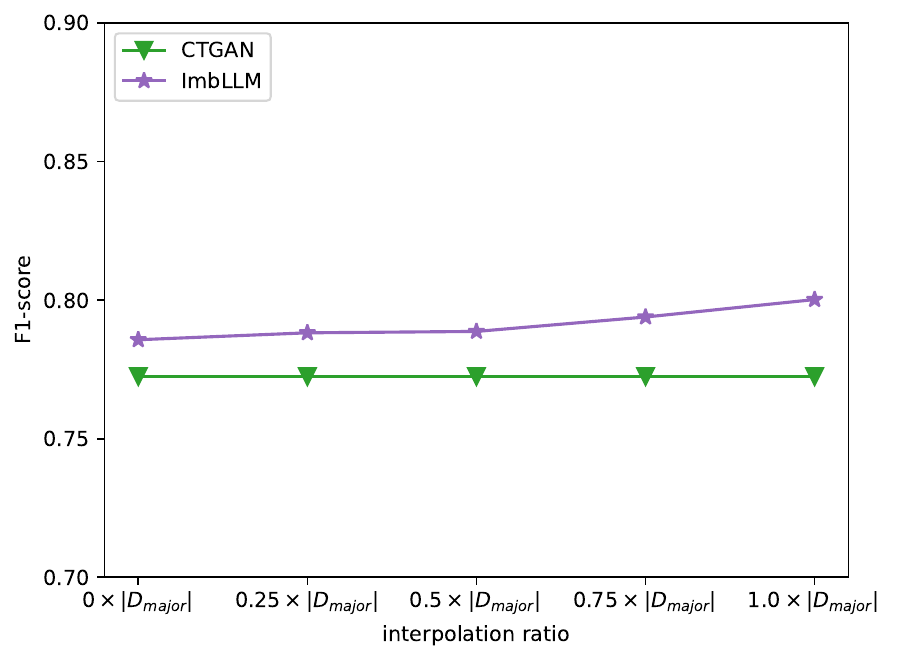}
\par\end{centering}
\caption{\label{fig:interpolation-ratio}F1-score vs. interpolation-ratio over
10 datasets.}
\end{figure}

Figure \ref{fig:interpolation-ratio} shows F1-scores of our method
ImbLLM with different values for $r$, compared with the best baseline
CTGAN. Note that at $r=0$, we have the performance of ImbLLM-inter
(the version of ImbLLM without interpolation) and at $r=1$, we have
the full performance of our method ImbLLM. We can see that our performance
increases when we set larger values for $r$, and our method is always
better than CTGAN with all values for $r$.

The results prove that our interpolation step is very effective to
generate diverse and generalizable synthetic minority samples, resulting
in better outcomes for imbalanced classification tasks.

\subsubsection{Effect investigation of imbalance-ratio}

Given the original set of minority samples ${\cal D}_{minor}^{*}$,
we use $\mid{\cal D}_{minor}\mid=q\times\mid{\cal D}_{minor}^{*}\mid$
as the number of minority samples in the training set of oversampling
methods. $q\in(0,1]$ is the imbalance-ratio, we set $q=0.2$ in our
experiments. In this study, we adjust $q$ to see whether it has impacts
on our performance.

\begin{figure}
\begin{centering}
\includegraphics[scale=0.5]{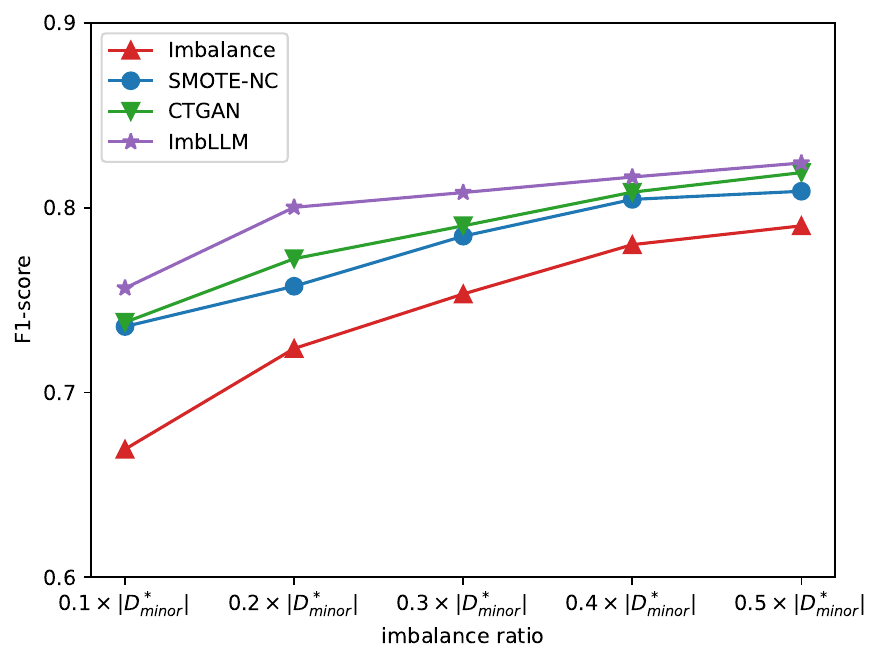}
\par\end{centering}
\caption{\label{fig:Train-size-Gen-size}F1-score vs. imbalance-ratio over
10 datasets.}
\end{figure}

Figure \ref{fig:Train-size-Gen-size} shows our method ImbLLM improves
when the number of minority samples increases. It improves significantly
when we enlarge $q$ from $0.1$ to $0.5$ (the improvement increases
from 0.7565 up to 0.8242). Other methods also benefit from more training
minority samples, e.g. Imbalance improves from 0.6693 up to 0.7902.
When the imbalance-ratio is severe (i.e. $q\in[0.1,0.3]$), there
is a big gap between our performance and those of the baselines. However,
the gap is smaller when more minority samples are presented in the
training set, as expected.

\subsubsection{Weakness of TapTap and Pred-LLM}

TapTap and Pred-LLM are two recent LLM methods that can effectively
generate synthetic minority samples. They involve an extra step to
verify and reject ill-generated minority samples. Namely, they use
a classifier/LLM trained/fine-tuned on the imbalanced dataset ${\cal D}_{train}$
to predict again the labels of synthetic minority samples. However,
this strategy can work well only in the case the imbalanced dataset
can represent the original dataset. In other words, if the classifier
trained on the imbalanced dataset (i.e. the Imbalance method) has
a poor performance, it may wrongly reject synthetic minority samples.
As shown in Table \ref{tab:TapTap's-weakness}, the Original method
trained with the original dataset has high F1-score. However, the
Imbalance method has performance reduced a lot on these four datasets,
leading to an ineffective verification step. As a result, TapTap and
Pred-LLM perform poorly. In contrast, as ImbLLM does not rely on the
verification step, it can reach to much better F1-scores.

\begin{table}[th]
\caption{\label{tab:TapTap's-weakness}TapTap and Pred-LLM weakness on four
datasets. Recall that they use a classifier/LLM trained/fine-tuned
on the imbalanced dataset to verify synthetic minority samples. However,
when the classifier does not perform well (indicated by the low F1-scores
of Imbalance), the verification step is ineffective, leading to poor
performance.}

\centering{}%
\begin{tabular}{|l|r|r|r|r|r|}
\hline 
\textbf{F1-score} & Original & Imbalance & TapTap & Pred-LLM & ImbLLM\tabularnewline
\hline 
\hline 
adult & 0.6761 & 0.4685 & 0.4705 & 0.5071 & \textbf{0.6543}\tabularnewline
\hline 
german & 0.5299 & 0.2032 & 0.2069 & 0.2363 & \textbf{0.4419}\tabularnewline
\hline 
bank & 0.3657 & 0.1118 & 0.0831 & 0.1352 & \textbf{0.3441}\tabularnewline
\hline 
sick & 0.9208 & 0.7790 & 0.7386 & 0.7740 & \textbf{0.8459}\tabularnewline
\hline 
\end{tabular}
\end{table}

\subsection{Generalization and diversity visualization}

We plot the \textit{distance to closest records} (DCR) histogram \cite{Borisov2023}
to show \textit{our synthetic minority samples are} \textit{similar
to }\textbf{\textit{unseen}}\textit{ minority samples in ${\cal D}_{test}$}.
The DCR metric computes the distance from a real minority sample to
its closest neighbor in the set of synthetic minority samples. Given
a real minority sample $x\in{\cal D}_{test}$, $\text{DCR}(x)=\text{min}\{d(x,\hat{x}_{i})\mid\hat{x}_{i}\in\hat{{\cal D}}_{minor}\}$,
where $d(\cdot)$ is an Euclidean distance.

\begin{figure*}[t]
\begin{centering}
\includegraphics[scale=0.5]{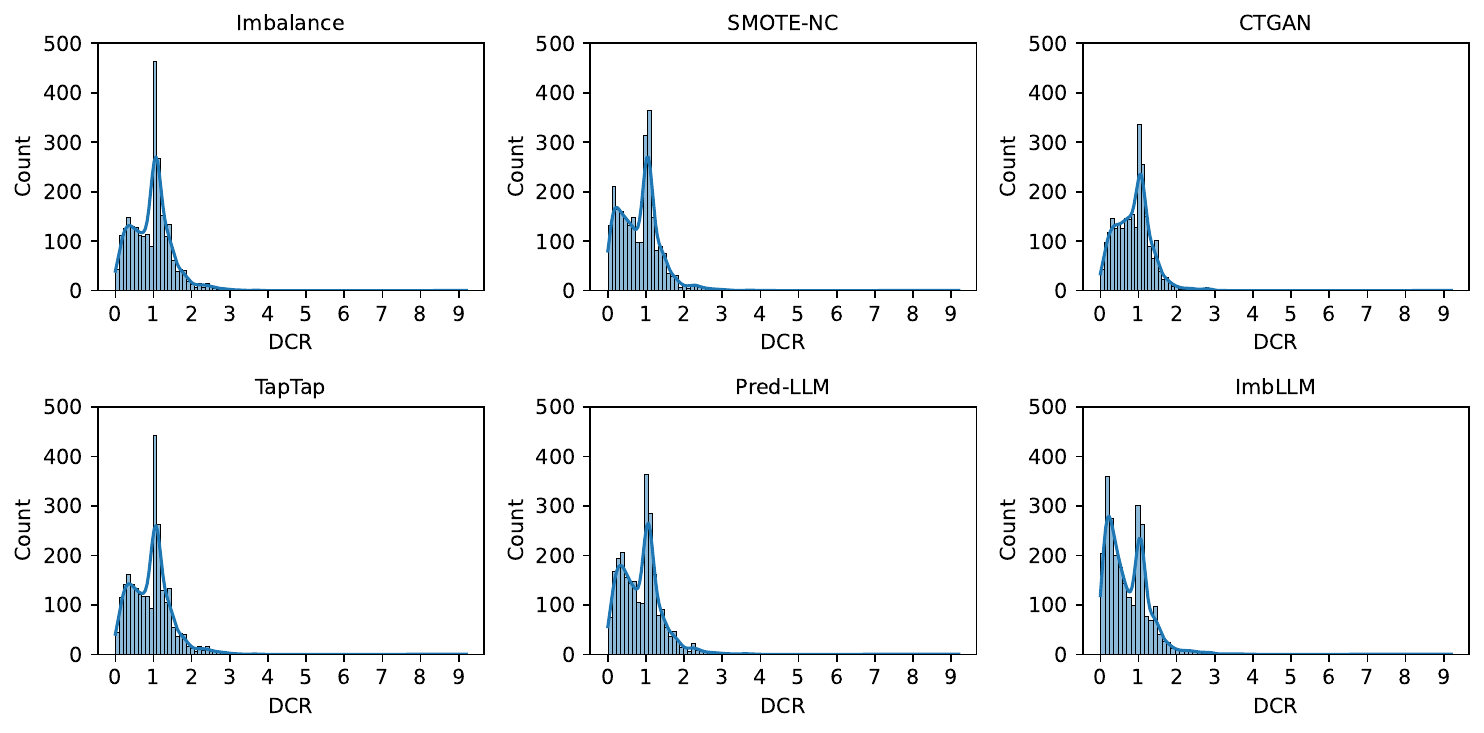}
\par\end{centering}
\caption{\label{fig:DCR-distributions}DCR distributions for the dataset \emph{gem\_price}.
While other methods show that their synthetic minority samples are
far away from the \textit{unseen} minority samples (their mode is
around 1), our ImbLLM shows that our synthetic minority samples are
close to the \textit{unseen} minority samples (our mode is around
0).}
\end{figure*}

Figure \ref{fig:DCR-distributions} visualizes the DCR distribution
of each method. Besides the top-3 methods, we also include the best
two LLM-based methods TapTap and Pred-LLM in the comparison list.
Only our method ImbLLM can generate synthetic minority samples in
close proximity to the \textit{unseen} minority samples. Other methods
show differences. Namely, most of our synthetic samples have the distances
to the unseen samples around $0$ while most of synthetic samples
of other methods have the distances to the unseen samples around $1$.
This result proves that our synthetic samples are diverse and generalizable
to cover many unseen samples.

\section{Related Works\label{sec:Related-Works}}

\subsection{Machine learning for tabular data}

Tabular data is one of the most popular data formats. In the ``AI
Report'' of Kaggle in 2023, it was estimated that between 50\% and
90\% of practicing data scientists used tabular data as their primary
data type in their experiments\footnote{https://www.kaggle.com/AI-Report-2023}.

ML and deep learning methods have been applied to tabular data in
many tasks. Some examples include: (1) \textit{table question answering}
answers a given question over tables \cite{Herzig2020}, (2) \textit{table
fact verification} checks if an assumption is valid for a given table
\cite{Chen2020}, (3) \textit{table to text} uses natural language
to describe a given table \cite{Bao2018}, and (4) \textit{table structure
understanding} identifies the table characteristics e.g. column types,
variable relations... \cite{Tang2021,Sui2024}.

Among these tasks, \textit{table generation} and \textit{table prediction}
(or \textit{table classification}) are the most well-known tasks used
in tabular data \cite{Assefa2020,borisov2022deep,Grinsztajn2022,hegselmann2023tabllm,Borisov2023,nguyen2024tabular}.

\subsection{Generative models for tabular data}

As generative models gained significant successes in image generation,
they have been adapted for \textit{table generation} (or \textit{tabular
data generation}) in different ways. Given a table, table generation
aims to learn a synthetic table that approximates the real table. 

Most tabular generation methods are based on Generative Adversarial
Network (GAN) \cite{Goodfellow2014} and Variational Autoencoder (VAE)
\cite{Kingma2019} models, including TableGAN \cite{Park2018}, CTGAN
\cite{Xu2019}, OCTGAN \cite{Kim2021}, and TVAE \cite{Xu2019}. Among
GAN-based methods, CTGAN \cite{Xu2019} is the most favored method
because it has three contributions to improve the data generation
process. First, it leverages different activation functions to generate
continuous and categorical data separately. Second, it normalizes
a continuous value based on its mode-specific instead of the mean
value or the min-max values of the corresponding column. Finally,
it uses a conditional generator to generate equal data for different
categories in categorical columns. These proposed steps greatly improve
the quality of the synthetic table.

As GAN-based methods require \textit{heavy pre-processing} steps,
LLM-based methods were introduced to address this problem \cite{Borisov2023,zhang2023generative,nguyen2024tabular}.
They show three advantages over GAN-based methods: (1) avoiding information
loss, (2) allowing context-aware, and (3) supporting arbitrary conditioning.
These benefits help them to generate realistic tabular data used in
various applications such as data augmentation \cite{seedat2024curated},
imbalanced classification \cite{yang2024language}, and few-shot classification
\cite{hegselmann2023tabllm}.

\subsection{Imbalanced classification in tabular data}

\textit{Table prediction} (or \textit{table classification}) is the
most popular task in tabular data. It predicts a label for the target
variable (e.g. ``\textit{Income}'') using a set of features (e.g.
``\textit{Age}'' and ``\textit{Education}''). While deep learning
methods are often dominant in other tasks, traditional ML methods
like XGBoost \cite{chen2016xgboost} still outperform deep learning
counterparts in table prediction.

\textit{Imbalanced classification} is a related task to table classification,
where the classes in the training table are not equally represented
(i.e. one class has a much larger samples than the others). There
are two main approaches to address the class-imbalance. Model-centric
approaches focus on modifying the objective functions in the ML classifiers
\cite{cao2019learning} or re-weighting the minority class \cite{cui2019class}.
A data-centric approach is oversampling, which generates more synthetic
minority samples.

\textit{Oversampling methods} are well-studied solutions for imbalanced
classification. Their principle is simple but effective. Most existing
methods rely on SMOTE (Synthetic Minority Oversampling Technology)
\cite{chawla2002smote}, where more minority samples are generated
by linearly combining two real minority samples to rebalance the dataset.
Several variants have been developed to address SMOTE weaknesses such
as outlier and noisy \cite{han2005borderline,sauglam2022novel}. Other
approaches are generative models such as CTGAN and TVAE \cite{Xu2019},
which learn the distribution of real minority samples via a generator
or encoder network.

Recently, LLMs have been adapted for oversampling methods \cite{Borisov2023,zhang2023generative,nguyen2024tabular,yang2024language}.
These methods often involve two main steps. First, they fine-tune
a pre-trained LLM with the imbalanced dataset. Then, they construct
prompts conditioned on the minority label to query the fine-tuned
LLM to generate minority samples. However, most of them focus on how
to re-verify the synthetic minority data but do not emphasize the
\textit{data diversity} that is an important factor in imbalanced
classification tasks.

\section{Conclusion\label{sec:Conclusion}}

LLM-based methods have emerged as potential solutions for oversampling
minority samples. However, they fail to produce diverse synthetic
data. To address this problem, we propose an LLM-based oversampling
method (named \textbf{ImbLLM}) with three important proposals in \textit{sampling},
\textit{permutation}, and \textit{fine-tuning}. Our method shows significant
improvements over eight SOTA baselines on 10 tabular datasets in imbalanced
classification tasks. It also shows our synthetic minority samples
have high quality and diversity based on qualitative and quantitative
metrics and visualization.

\balance

\bibliographystyle{plain}
\bibliography{reference}

\end{document}